\documentclass[conference]{IEEEtran}
\IEEEoverridecommandlockouts
\usepackage{cite}
\usepackage{amsmath,amssymb,amsfonts}
\usepackage{algorithmic}
\usepackage{graphicx}
\usepackage{textcomp}
\usepackage{xcolor}
\usepackage{amsmath}
\def\BibTeX{{\rm B\kern-.05em{\sc i\kern-.025em b}\kern-.08em
    T\kern-.1667em\lower.7ex\hbox{E}\kern-.125emX}} 

\begin{document}
\title{\LARGE\bf
GameVLM: A Decision-making Framework for Robotic Task Planning Based on Visual Language Models and Zero-sum Games}

\author{\IEEEauthorblockN{Aoran Mei, Jianhua Wang, Guo-Niu Zhu*, Zhongxue Gan*}
\IEEEauthorblockA{\textit{Academy for Engineering and Technology, Fudan University, Shanghai 200433, China}\\
\textit{*Corresponding author: guoniu\_zhu@fudan.edu.cn, ganzhongxue@fudan.edu.cn}}}

\maketitle

%%%%%%%%%%%%%%%%%%%%%%%%%%%%%%%%%%%%%%%%%%%%%%%%%%%%%%%%%%%%%%%%%%%%%%%%%%%%%%%%
\begin{abstract}
With their prominent scene understanding and reasoning capabilities, pre-trained visual-language models (VLMs) such as GPT-4V have attracted increasing attention in robotic task planning. Compared with traditional task planning strategies, VLMs are strong in multimodal information parsing and code generation and show remarkable efficiency. Although VLMs demonstrate great potential in robotic task planning, they suffer from challenges like hallucination, semantic complexity, and limited context. To handle such issues, this paper proposes a multi-agent framework, i.e., GameVLM, to enhance the decision-making process in robotic task planning. In this study, VLM-based decision and expert agents are presented to conduct the task planning. Specifically, decision agents are used to plan the task, and the expert agent is employed to evaluate these task plans. Zero-sum game theory is introduced to resolve inconsistencies among different agents and determine the optimal solution. Experimental results on real robots demonstrate the efficacy of the proposed framework, with an average success rate of 83.3\%. Videos of our experiments are available at \textsl{\textcolor[rgb]{0,0,1}{https://youtu.be/sam-MKCPP7Y}}.
\end{abstract}


%%%%%%%%%%%%%%%%%%%%%%%%%%%%%%%%%%%%%%%%%%%%%%%%%%%%%%%%%%%%%%%%%%%%%%%%%%%%%%%%
\section{Introduction}
Classical rule-driven and learning-based planning algorithms have been widespread in robotic task planning. However, the rule-driven framework requires extensive domain knowledge. It faces challenges due to the strong assumptions of perfect perception and action executions \cite{zhang2023grounding}. Although rule-based systems perform well in handling predefined tasks under structured environments, their adaptability and flexibility would be challenged when encountering unknown factors in unstructured or dynamic scenarios \cite{alterovitz2016robot}. When faced with novel tasks or dynamic environments, these rule-based systems may not effectively manage unforeseen situations. Due to a lack of scene perception and understanding, robots struggle to adjust and optimize their actions when executing complex task planning. As a result, the efficiency and accuracy of task execution would be decreased. On the other hand, learning approaches, such as hierarchical reinforcement learning and imitation learning, often require intricate reward engineering and time-consuming dataset creation efforts \cite{skreta2024replan}.

In contrast, the emergence of vision-language models (VLMs) has shown considerable promise in robotic task planning \cite{driess2023palm, brohan2023rt, bao2023smart}. By combining advanced computer vision technology with natural language processing techniques, these VLMs enable robots to identify various objects and their attributes more accurately while understanding complex commands and environmental contexts during task execution \cite{ghosh2024exploring, shah2023lm}. With VLMs, robots can not only recognize the basic shapes and sizes of objects but also understand their affordances and their spatial relationships with other objects \cite{wang2024exploring}. Based on their sophisticated scene understanding and reasoning abilities, VLMs can effectively advance the intelligence and automation of robotic task planning and execution \cite{hu2023look, wake2023gpt}. 

However, VLMs also struggle with hallucinations, the complexity of semantics, and limited context. They may generate incorrect responses even when the inputs are accurate \cite{liu2024survey}. To resolve such issues, multi-agent strategies were investigated in past studies to reduce the probability of generating incorrect decisions \cite{xi2023rise, melnic2023multi}. Nevertheless, the introduction of multiple agents poses new challenges. How do we deal with the inconsistency in the outputs from different agents?

\begin{figure}[ht]
\centering
\includegraphics[width=8cm]{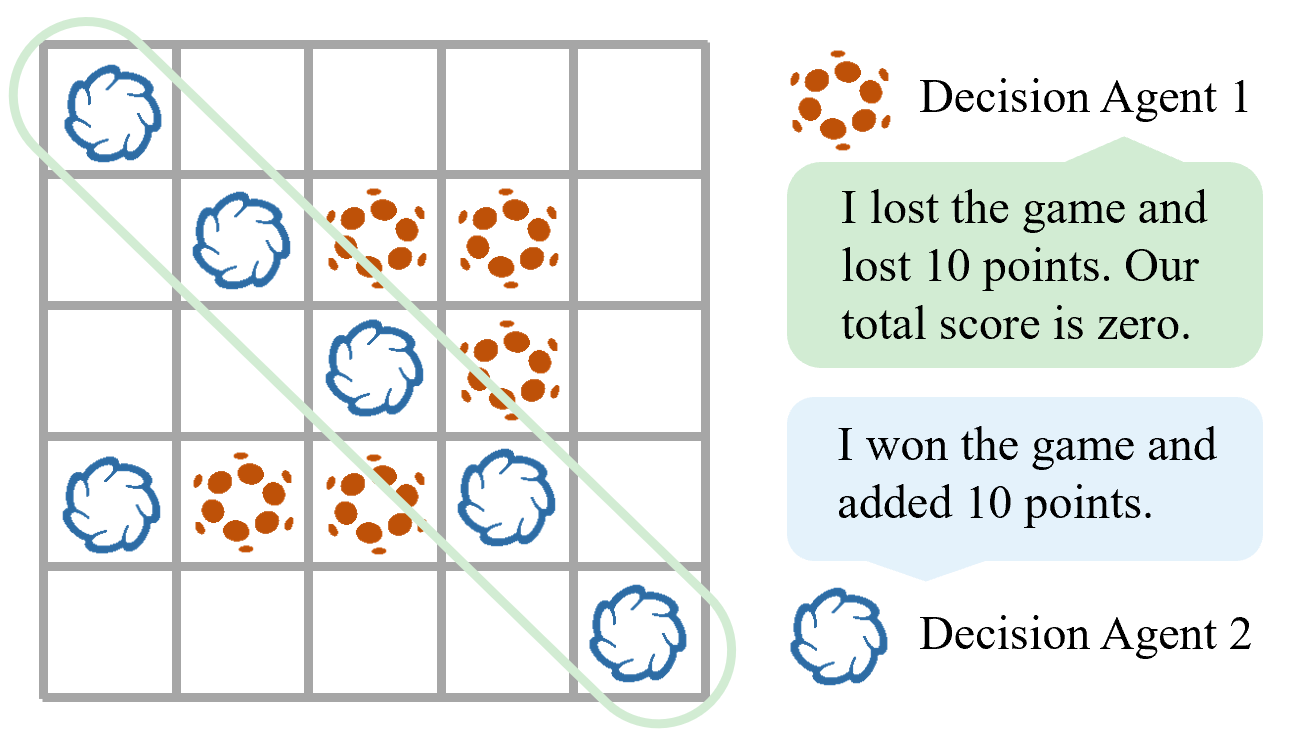}
\caption{An example of zero-sum game theory and multi-agents: the game of Gomoku. Each player is an agent. The winning player gains ten points, while the losing player loses ten points. The total sum of their scores remains constant.}
\label{game}
\end{figure}

Inspired by the game theory in group decision-making, this study aims to investigate the zero-sum game approach to resolve the inconsistency between these agents. As a well-known concept in game theory, the zero-sum game shows great strength in clarity and predictability. Its strategic insight makes it an invaluable tool in areas where competitive interactions are prevalent \cite{hwang2020strategic}. As shown in Fig. \ref{game}, an interaction framework is constructed based on a question-and-answer mechanism. Each agent challenges others' strategies by asking questions and must answer the questions raised by others. An expert-level agent is introduced to evaluate the questions and answers. The respondent loses ten points for each inconsistent response, while the questioner will be awarded ten points. After several rounds of questioning and answering, the strategy given by the agent with the highest score will be selected as the final.

In view of the merit of VLMs in scene understanding and the strength of zero-sum game theory in multi-agent decision-making, this paper proposes a robotic task planning framework, namely GameVLM. Multiple decision agents are introduced to generate respective task plans. Then, an expert agent is developed to evaluate the consistency of these task plans. A zero-sum game approach is presented to determine the optimal strategy. A real-time object detection model, YOLO-World, is used to identify the object's coordinates for robotic grasping. 

The main contributions of this paper are summarized as follows.
\begin{itemize}
\item This study proposes a GameVLM, a robotic task planning framework based on VLMs and zero-sum game theory. Multiple agents are used to generate task plans, while an expert agent is adopted to evaluate these plans.
\item Zero-sum game theory is introduced to resolve the inconsistency between different agents. The final solution is obtained through a question-and-answer mechanism.
\item Multiple tasks are presented to evaluate the applicability of the proposed method. Experimental results on real robots demonstrate the efficacy of the proposed GameVLM.
\end{itemize}

\section{Methodology}
In this section, the structure and functionality of the GameVLM framework are discussed in detail. The components of the GameVLM, including the input module, decision agent, expert agent, and object detection module, are outlined.

\subsection{Framework of the Proposed GameVLM}
\begin{figure*}[ht]
\centering
\includegraphics[width=17.5cm]{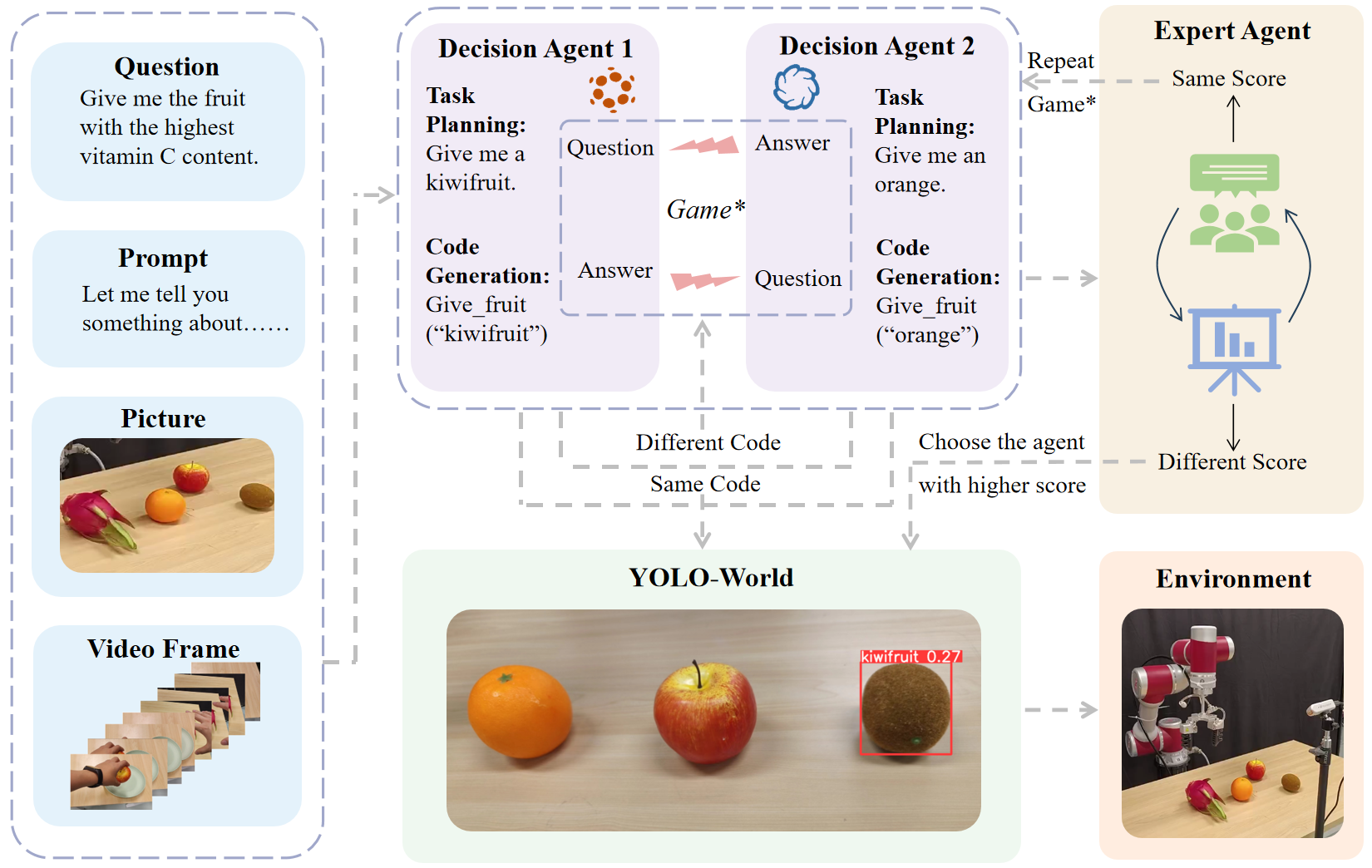}
\caption{GameVLM overview. We propose a GameVLM framework, which comprises an input module, two decision agents, an expert agent, and an object detection module. The decision and expert agents refer to VLMs. Two decision agents are used to generate task plans and codes, while an expert agent checks the consistency of these codes. A real-time open-vocabulary object detection model, i.e., YOLO-World, is introduced to detect objects in the image.}
\label{fig-overall}
\end{figure*}

By integrating VLMs and zero-sum game theory, the GameVLM framework is proposed to enhance decision-making in robotic task planning. As illustrated in Fig. \ref{fig-overall}, the GameVLM framework consists of an input module, two decision agents, an expert agent, and an object detection module. First of all, the GameVLM framework takes text and visual information, such as prompts, images, and video frames, as input. Then, the input information is passed to decision agents, which are responsible for task planning and code generation. The initial state of each decision agent is set to zero. After that, an expert agent is introduced to check the consistency of the codes generated by the decision agents. If the codes are consistent, they will be passed to the robot for execution. Otherwise, the decision agents will be asked to engage in three rounds of interaction through questions and answers to get a compromise solution. For each unsatisfactory answer, the respondent will be deducted ten points, while the questioner obtains ten points. By contrast, the respondent gets ten points, and the questioner loses ten points for each satisfactory answer. Based on such evaluation mechanisms, the robot will execute the code generated by the decision agent with higher scores. With such a zero-sum game strategy, the GameVLM is able to promote communication between decision agents and enhance the accuracy and consistency of decisions through competition and collaboration.

\subsection{Input Module}
In this module, the input information can be taken in various modalities, such as text, image, and video frames. For example, in a text description like ``pick up an apple," the input text directly specifies the required action. For commands such as ``arrange the blocks in the order as shown in the picture," the module uses text and visual information as input to facilitate the planning and execution of complex tasks. For video input, such as a set of video frames, instructions can be described as ``predict and perform the next action in the video." The module takes text and video information as input. 

By taking multi-modal information as input, the clarity of the instruction and the flexibility and accuracy of task completion can be effectively enhanced.

\subsection{Decision Agent}
There are two agents in the decision module. Each decision agent consists of a VLM model, i.e., GPT-4V, and its prompts. The decision agent is responsible for processing information from the input module and generating task plans and executable codes. If the code generated by one agent is inconsistent with the one from another agent, it will challenge another agent by posing questions. Correspondingly, it should respond to inquiries from its partner. With such an interaction mechanism, the decision-making process can be optimized, and a compromise solution can be obtained.

\subsection{Expert Agent}
The expert agent is used to check the consistency of the codes generated by the decision agents. If the codes generated by the two decision agents are consistent, they will be sent to the robot for execution. Otherwise, the expert agent will ask the decision agents to compete with each other through mutual questioning. The respondent will be deducted ten points for each wrong answer. By contrast, the questioner gains ten points. After three rounds of questioning and answering, the codes generated by the agent with the higher score will be identified as the final one and sent for execution. If there is a tie in scores, the questioning and answering process will be repeated until a score difference emerges.

\subsection{Object Detection} 
\begin{figure}[ht]
\centering
\includegraphics[width=8cm]{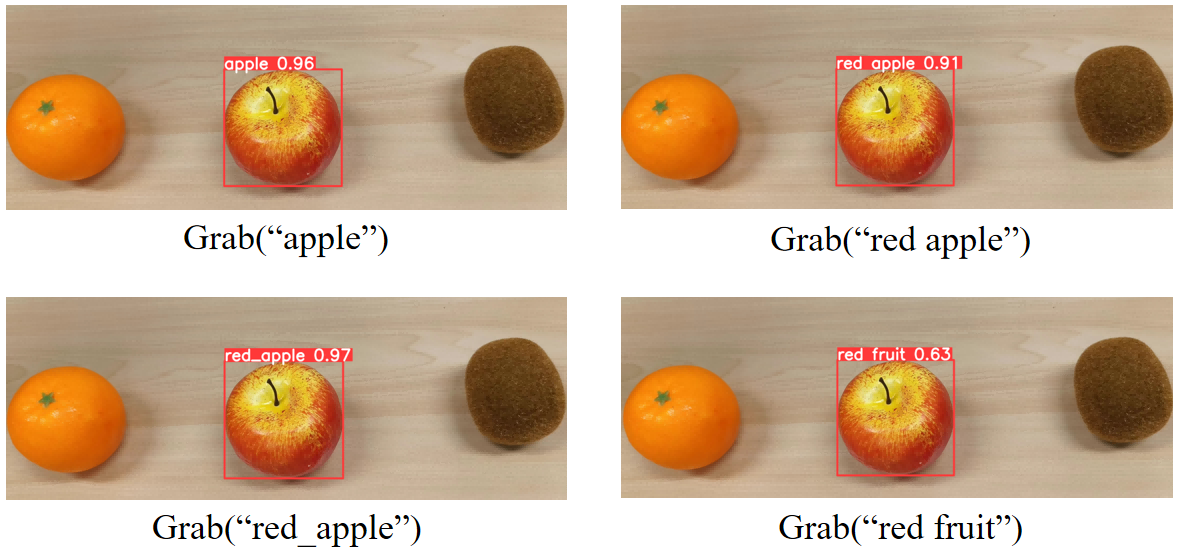}
\caption{Demonstration of the YOLO-World on real images. The model can identify apples in the image, regardless of whether they are labeled as ``apple," ``red apple," ``red\_apple," or ``red fruit."}
\label{fig-yolo}
\end{figure}

As depicted in Fig. \ref{fig-overall}, a real-time open-vocabulary object detection model, i.e., YOLO-World \cite{cheng2024yolo}, is introduced to detect objects in the image. YOLO-World is a zero-shot model that can precisely identify objects and obtain their pixel coordinates in RGB images without pre-training. As shown in Fig. \ref{fig-yolo}, YOLO-World shows a high robustness towards object names. It can identify the apple in the image, whether it is referred to as “apple,” “red\_apple,” or “red apple.” Based on the flexibility of the YOLO-World, the tolerance and robustness of the GameVLM framework are effectively enhanced.

\section{Experiments}
This section presents the protocol of real-world experiments conducted to evaluate the efficacy of the proposed GameVLM.

\subsection{Experimental Setup}
\begin{figure}[ht]
\centering
\includegraphics[width=8cm]{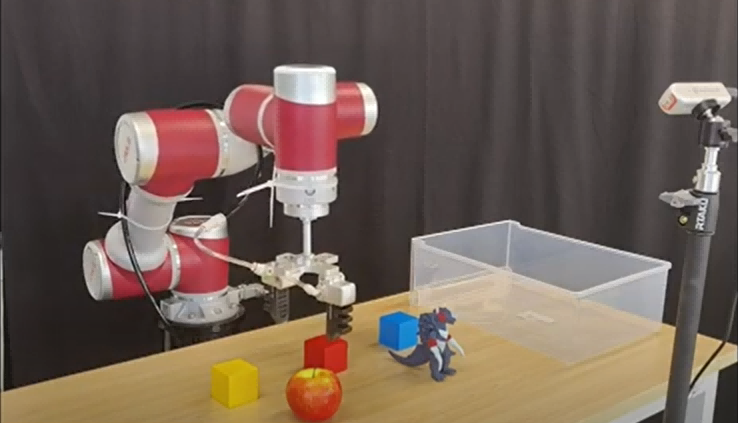}
\caption{Overview of the system setup.}
\label{setup}
\end{figure}

As illustrated in Fig. \ref{setup}, the experimental setup comprises a six-degrees-of-freedom robotic arm (i.e., JAKA Zu 7), a Rochu pneumatic gripper, and an Intel Realsense D435i depth camera. Some simulated fruits, blocks, and monster models are introduced as experimental subjects. During the experiment, the robotic arm is commanded to grasp specific objects according to the results of the visual perception.

\subsection{Prompt}
\begin{figure*}[ht]
\centering
\includegraphics[width=17.5cm]{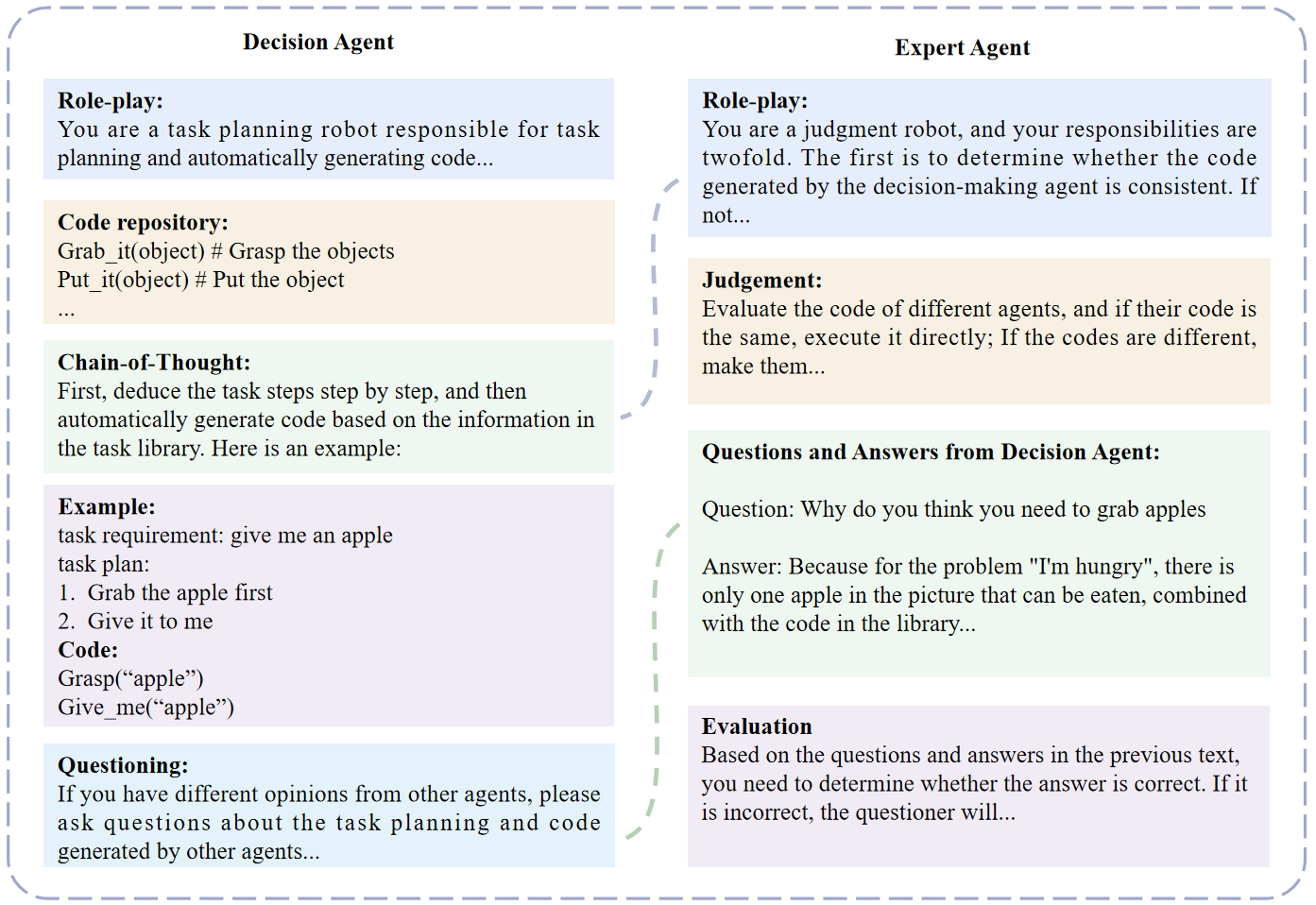}
\caption{Example prompts in the GameVLM framework.}
\label{prompt}
\end{figure*}

As depicted in Fig. \ref{prompt}, the prompt of the decision agent comprises five modules: role-playing, code repository, chain-of-thought\cite{wei2022chain}, examples, and questions. These modules work together to enable the agent to handle intricate situations effectively. Similarly, the prompt of the expert agent consists of four modules: role-playing, judgment, question and answer, and evaluation. These modules are integrated to empower the agent to identify and resolve problems accurately. 

\subsection{Task Description}
A couple of tasks with various characteristics are proposed to evaluate the efficacy of the proposed GameVLM in real-world experiments. As listed in Table \ref{table-1}, these tasks are presented based on the following criteria.

\begin{table}[htbp]
\caption{Characteristics of the experimental task}
\begin{center}
\begin{tabular}{|c|c|c|c|c|c|c|} 
\hline
Task & SU & SR & SUG & VU & WKU & FP \\
\hline
Task 1 &  &  & \checkmark &  & \checkmark & \\
\hline
Task 2 & \checkmark &  & \checkmark &  &  & \\
\hline
Task 3 & \checkmark & \checkmark & \checkmark &  &  & \\
\hline
Task 4 &  & \checkmark & \checkmark &  &  & \\
\hline
Task 5 &  &  & \checkmark & \checkmark &  & \\
\hline
Task 6 &  & \checkmark & \checkmark & \checkmark & \checkmark & \checkmark\\
\hline
\end{tabular}
\label{table-1}
\end{center}
\end{table}

\textbf{Semantic understanding (SU):} the ability to filter redundant information and grasp the underlying meanings. For example, for the command ``I'm hungry," the system can derive that the user needs an apple based on the environmental information.

\textbf{Spatial reasoning (SR):} the ability to identify spatial relationships, such as the vertical arrangement of objects. 

\textbf{Scene understanding: (SUG):} the ability to understand the information in a scene, such as identifying constraints in a specific setting.

\textbf{Video understanding (VU):} the ability to understand information based on a sequence of video frames, such as replicating the actions shown in a video.

\textbf{World knowledge understanding (WKU):} the ability to apply known knowledge to complete tasks, such as determining whether a cartoon character is evil or good.

\textbf{Future prediction (FP):} the ability to predict future actions based on the current state of the scene. 

Based on these criteria, the experimental tasks are proposed as follows.

\textbf{Task 1: Grasp special objects. } There are apples, kiwifruit, and blocks on a desk. This task requires the robot to understand that kiwifruit has the highest vitamin C content and grasp the kiwifruit on the desk.

\textbf{Task 2: Organize the desktop.} There are blocks, monster toys, and a storage box on a desk. The robot needs to place the blocks and monster toys in the storage box.  

\textbf{Task 3: Place the block.} In this task, we do not specify where to place the block. Instead, we provide a picture pointing to a pink plate. Correspondingly, the robot is required to place the block on the pink plate.

\textbf{Task 4: Stack blocks.} By providing a picture of a yellow block stacked on top of a red block, the robot needs to stack blocks similarly.

\textbf{Task 5: Imitate the behavior. } This task requires the robot to understand the action of placing an apple on a plate based on the information in the video frames and then replicate this behavior.

\textbf{Task 6: Predict the next action.} In this task, the robot is required to predict and perform the action of grasping an orange based on the human's movement in the video.

\begin{figure*}[ht]
\centering
\includegraphics[width=17.5cm]{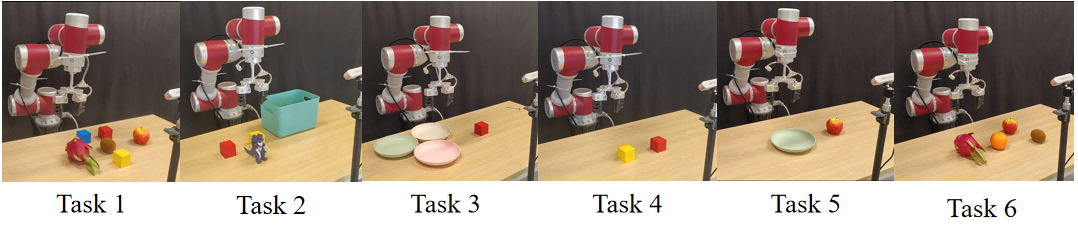}
\caption{Experimental setup in the real environment.}
\label{fig-task_setup}
\end{figure*}

The experimental setup is illustrated in Fig. \ref{fig-task_setup}. In the experiment, each task will be run in ten rounds. The average success rate of the ten rounds is taken as the evaluation metric to examine the performance of the GameVLM in real-world scenarios. Moreover, Table \ref{table_2} lists the difficulty levels of these six tasks and the instructions that need to be entered for each task. The difficulty level is defined as follows.

\textbf{Low:} In the task planning process, the agent needs to comprehend the implicit meanings within the instructions.

\textbf{Middle:} Besides the low-difficulty level capabilities, the agent also needs to comprehend the implicit meanings within images or video frames.

\textbf{High:} Other than the middle-difficulty level abilities, the agent also needs to predict possible future behaviors.

\begin{table}[ht]
\caption{Characteristics of the experimental task}
\begin{center}
\begin{tabular}{|c|c|c|} 

\hline
Method & Difficulty & Instruction \\
\hline
Task 1 & Low &  The object with the highest vitamin C content.  \\
\hline
Task 2 & Middle &  Organize the table. \\
\hline
Task 3 & Middle & Place the block in that location.   \\
\hline
Task 4 & Low & Stack blocks as shown in the picture.   \\
\hline
Task 5 & Low &  Imitate the behavior in the video.  \\
\hline
Task 6 & High & Predict the next action in the video. \\
\hline
\end{tabular}
\label{table_2}
\end{center}
\end{table}

\section{Results}
Corresponding to the experiment protocol, the experimental results are presented.

\subsection{Experimental Results: From a Task Perspective}
Table \ref{table_3} lists the experimental results from a task perspective. As shown in Table \ref{table_3}, the GameVLM framework gets an average success rate of 83.2\% for all six tasks. The success rate of the six tasks are obtained as 90\% (task 1), 80\% (task 2), 80\% (task 3), 90\% (task 4), 100\% (task 5), and 60\% (task 6). Among these tasks, the GameVLM framework achieves the highest success rate (100\%) in task 5 (imitate the behavior). On the contrary, it gets the lowest success rate (60\%) in task 6 (predict the next action). In most of the tasks, the success rate can reach up to 80\% or even higher. Overall, the GameVLM performs well across various tasks, especially in grasping special objects (task 1), stacking objects (task 4), and imitating the behavior (task 5). The success rate of these three tasks reaches up to 90\% or even 100\%.

\begin{table*}[ht]
\caption{Results of the experiment: from a task perspective}
\begin{center}
\begin{tabular}{|c|c|c|c|c|c|c|c|} 
\hline
Task &Task 1 & Task 2 & Task 3 & Task 4 & Task 5 & Task 6 & Average\\
\hline
Success rate & 90\% & 80\% & 80\% & 90\%  &  100\% &  60\% & 83.2\%  \\
\hline
\end{tabular}
\label{table_3}
\end{center}
\end{table*}

From the experimental results, the GameVLM shows outstanding performance in imitating the behavior in the video. The system can effectively understand the actions shown in the video and replicate these behaviors. It also achieves good results in grasping special objects and stacking blocks. In these two tasks, the system illustrates prominent ability in scene understanding and reasoning. By contrast, the GameVLM performs worst in predicting the next action in the video. It is challenging for the system to predict and perform the next action. Moreover, the success rate of these six tasks corresponds to their difficulties. The system performs better on low-difficulty tasks and less on high-difficulty tasks.

In general, the GameVLM performs well in handling a variety of real-world tasks, especially those involving visual and spatial understanding and semantic logic processing. However, the system needs further enhancement to interpret and predict future events based on current cues.

\subsection{Experimental Results: From a Criteria Perspective}
Table \ref{table_4} presents the experimental results from a criteria perspective. As depicted in Table \ref{table_4}, the success rate of GameVLM on the six criteria is listed as 80\% (SU), 76.7\% (SR), 83.3\% (SUG), 80\% (VU), 75\% (WKU), and 60\% (FP). Among these criteria, the GameVLM obtains the highest success rate on SUG (scene understanding), with a success rate of 83.3\%. On the contrary, the GameVLM gets the lowest success rate on FP (future prediction), with a success rate of 60\%. In most of the criteria, the success rate is higher than 75\%.
\begin{table}[ht]
\caption{Results of the experiment: from a criteria perspective}
\begin{center}
\begin{tabular}{|c|c|c|c|c|c|c|} 
\hline
Method &SU & SR & SUG & VU & WKU & FP \\
\hline
Success rate & 80\% & 76.7\% & 83.3\% & 80\%  & 75\% &  60\%   \\
\hline
\end{tabular}
\label{table_4}
\end{center}
\end{table}

As shown in Table \ref{table_4}, the GameVLM gets a success rate of 80\% or even higher on SU (semantic understanding), SUG (scene understanding), and VU (video understanding). It gets a success rate of 76.7\% on SR (spatial reasoning) and 75\% on WKU (world knowledge understanding). From the experimental results, the GameVLM shows strong ability in various understanding scenarios, such as semantic understanding, scene understanding, and video understanding. The system can understand the information in a scene, and the information based on a sequence of video frames. It can also grasp the underlying meanings from the semantic description. Besides these criteria, the GameVLM performs well on spatial reasoning and world knowledge understanding. The system is able to identify spatial relationships and apply known knowledge to complete tasks. In contrast, the GamevLM performs worst in future prediction. It is a challenge for the system to predict future actions based on the scene's current state.

\section{Conclusion}
To enhance the success rate of robots in task planning and execution, this paper proposed a GameVLM framework, which integrates VLMs and zero-sum game theory. In this framework, a multi-agent strategy was adopted to increase the accuracy of the decision-making in robotic task planning. Two decision agents were presented to plan tasks and generate codes. An expert agent was introduced to evaluate the consistency of the codes generated by these two decision agents. Zero-sum game theory was used to get an optimal solution. A couple of real-world experiments were conducted to evaluate the efficacy of the proposed GameVLM. Experimental results on real robots demonstrate the superiority of the proposed framework. By integrating VLMs, multi-agent, and zero-sum game theory, the GameVLM framework performs well in various tasks. It not only enhances the flexibility of task planning but also improves the robustness of the robotic system during task execution.

Although the GameVLM works well on scene understanding and reasoning, it faces challenges in prediction. In the future, more efforts will be devoted to investigating the planning of long-term tasks.

\section*{Acknowledgement}
This work is supported by Shanghai Municipal Science and Technology Major Project under Grant 2021SHZDZX0103 and Key Project of Comprehensive Prosperity Plan of Fudan University under Grant XM06231744.

%%%%%%%%%%%%%%%%%%%%%%%%%%%%%%%%%%%%%%%%%%%%%%%%%%%%%%%%%%%%%%%%%%%%%%%%%%%%%%%%
\bibliographystyle{IEEEtran}
\bibliography{ref}
\end{document}